\documentclass{article}
\usepackage{spconf,amsmath,graphicx}
\usepackage[table]{xcolor}
\usepackage{hyperref}
\usepackage{booktabs}
\usepackage{pgf-pie}
\usepackage{multirow}

\definecolor{lgray}{gray}{0.95}

\title{Sens-VisualNews: A Benchmark Dataset for Sensational Image Detection}

\name{Andreas Goulas$^{\star,\dagger}$, Damianos Galanopoulos$^{\star}$, Evlampios Apostolidis$^{\star}$, Vasileios Mezaris$^{\star}$\thanks{This work was supported by the EU’s Horizon Europe programme under grant agreement 101070190 AI4Trust.}}
\address{$^{\star}$Information Technologies Institute (ITI), CERTH, Thessaloniki, Greece \\
$^{\dagger}$Queen Mary University of London, London, UK \\
\{agoulas, dgalanop, apostolid, bmezaris\}@iti.gr}

\begin{document}
%
\maketitle
\begin{abstract}
The detection of sensational content in media items can be a critical filtering mechanism for identifying check-worthy content and flagging potential disinformation, since such content triggers physiological arousal that often bypasses critical evaluation and accelerates viral sharing. In this paper we introduce the task of sensational image detection, which aims to determine whether an image contains shocking, provocative, or emotionally charged features to grab attention and trigger strong emotional responses. To support research on this task, we create a new benchmark dataset (called Sens-VisualNews) that contains 9,576 images from news items, annotated based on the (in-)existence of various sensational concepts and events in their visual content. Finally, using Sens-VisualNews, we study the prompt sensitivity, performance and robustness of a wide range of open SotA Multimodal LLMs, across both zero-shot and fine-tuned settings.
\end{abstract}

\section{Introduction}
\label{sec:intro}

Several recent works point out the tendency of disinformation to adopt a sensationalist story format \cite{13020024, HAMBY2024114289, SUI2023107654}. So, the detection of sensational content seems to be essential for spotting news items that require fact-checking. For this, several approaches have been described to identify media that uses provocative, exaggerated, or emotionally charged language to grab attention, e.g., \cite{HAMBY2024114289,10549819,Alarfaj2025}.
Nevertheless, disinformation can also include the use of sensational images that aim to grab immediate attention, provoke intense emotional reactions, or manipulate user perception through dramatic or shocking imagery. Such visual content serves as a potent catalyst for spreading disinformation, since its ability to trigger physiological arousal often bypasses critical evaluation and accelerates content sharing. Hence, there is a need for technologies for sensational image detection, that could be used as a filtering mechanism and assist fact-checkers to identify check-worthy content and flag potential disinformation.

As stated above, existing methods for sensational content detection analyze textual content aiming to identify emotionally charged language or phrases designed to evoke fear, anger or extreme curiosity \cite{HAMBY2024114289,10549819,Alarfaj2025}. Two different classes of methods that focus on the impact of visual content on humans, are the ones dealing with NSFW (Not Safe For Work) content detection, and visual sentiment analysis. The methods of the first class aim to spot visual material that is inappropriate for public, professional or general viewing (e.g., visually disturbing content) or violates safety guidelines \cite{zhang2025usd,Chandra_2025,10769133}. The methods of the second class aim to identify the emotional tone or attitude expressed in the visual content \cite{jiang2024,WANG2025109731,10.1145/3709147}. Hence, such methods can support the detection of sensational images for flagging disinformation, only to a limited extent.

To fill this gap, in this paper, we introduce the task of sensational image detection, which aims to determine whether an image contains shocking, provocative, or emotionally charged features to grab attention and trigger strong emotional responses (e.g., shock, fear, anger, disgust, anxiety). Following, we employ the VisualNews \cite{liu2021visual} large-scale dataset for news image captioning, and annotate a subset of images (9,576 in total) according to the (in-)existence of various sensational visual concepts and events that appear in disinformation items, formulating a new benchmark dataset (called Sens-VisualNews) for sensational image detection. Using the created dataset, we investigate the performance of several SotA Multimodal Large Language Models (MLLMs) on the proposed task. Our contributions are as follows:
\begin{itemize}
    \item We introduce the task of sensational image detection that deals with the identification of images that aim to trigger strong emotional responses.
    \item We create the Sens-VisualNews\footnote{Our dataset annotations and code are available at \url{https://github.com/IDT-ITI/Sens-VisualNews}} benchmark dataset with 9,576 annotated images based on the (in-)existence of various sensational visual concepts and events.
    \item Using Sens-VisualNews, we study the performance of families of open SotA Multimodal LLMs on this task, across both zero-shot and fine-tuned settings.
\end{itemize}

\section{Related Work}
\label{sec:literature}

The research domain on \textbf{sensational content detection},
deals with the development of methods for identifying news, social media posts, or digital content designed to provoke intense emotions (e.g., fear, anger, shock), often to maximize clicks or engagement. So, it \textbf{focuses on the analysis of textual content}, using NLP and machine or deep learning methodologies. For example, Hamby et al., \cite{HAMBY2024114289} investigated the role of different narrative characteristics in identifying sensational stories and predicting the spread of disinformation. Wang et al., \cite{10549819} described a clickbait detection method that uses prompt-tuning for leveraging the few-shot labeled titles during the training of the detector.
Finally, Alarfaj et al., \cite{Alarfaj2025} presented a transformer-based network architecture for automated detection of clickbait news headlines, and compared its performance against SotA machine- and deep-learning approaches from the literature, using a benchmarking dataset.

With regards to the emotional impact of visual content on humans, this is assessed by methods from two different research domains. The methods for \textbf{NSFW content detection} aim to identify \textbf{visual content that is inappropriate for public, professional or general viewing} (e.g., visually disturbing content) or \textbf{violates safety guidelines}. For example,
Zhang et al., \cite{zhang2025usd} presented an approach that leverages scene graph generation and classification to detect harmful attributes and relationships within images.
Chandra et al., \cite{Chandra_2025} investigated the performance of CNN and VGG-16 network architectures in identifying pornographic content. Finally, Tzelepi et al., \cite{10769133} represented the generic semantic descriptions and elicited emotions encoded in LMMs using CLIP-based representations, and combined the obtained representations with CLIP-based image embeddings for performing visually-disturbing image detection.

From a different standpoint, the methods for \textbf{visual sentiment analysis} try to identify the \textbf{emotional tone or attitude expressed in the visual content}. For example, Jiang et al., \cite{jiang2024} described a network architecture for visual sentiment analysis, comprising a fine-tuned VGG16 model using SVM with augmented training data from two Twitter image datasets. 
Wang et al., \cite{WANG2025109731} presented a method for multimodal sentiment analysis, that takes into account the internal correlation between sentiment-related representations from textual and visual data. On the same direction, Mu et al., \cite{10.1145/3709147} fine-tuned the BLIP-2 Vision-Language Model (VLM) with LoRA, to align and adapt the learned representations for the needs of multimodal sentiment prediction.

Our literature review demonstrates the limited capacity of existing methods from the domains of sensational content detection, NSFW content detection and visual sentiment analysis, to support the detection of sensational images for flagging disinformation. This observation motivated us to introduce the task of sensational image detection and propose a new benchmark for performance comparison. 

\section{Proposed benchmark}

\subsection{Problem statement}

The task of sensational image detection aims to identify the existence of shocking, provocative or emotionally charged visual features, intended to grab the viewers' attention and trigger strong emotional responses (e.g., shock, fear, anger, disgust and anxiety). It differs from the task of sensational content detection, which focuses on the detection of provocative, exaggerated or emotionally charged language. Moreover, it has a different and broader scope compared to the task of NSFW visual content detection, which focuses on the identification of harmful material that is inappropriate for public, professional or general viewing. Finally, it deviates from the task of visual sentiment analysis, as it focuses on identifying if an image is used to trigger strong emotional responses, rather than estimating  the emotional tone or general polarity evoked by an image. The development of methods for sensational image detection and their integration into journalistic and news applications, will provide journalists and fact-checkers the means to spot check-worthy content and flag potential disinformation before it gets viral.

\subsection{Source dataset}

The basis for building our benchmark was the VisualNews dataset for news image captioning \cite{liu2021visual}, which contains over one million news images, along with articles, captions and other metadata from four news agencies (``The Guardian'', ``BBC'', ``USA Today'', ``Washington Post''). Besides its use by researchers working on news image captioning, this dataset has been extended and re-purposed to support other tasks, such as out-of-context multimodal data detection \cite{luo-etal-2021-newsclippings}. Its visual content is highly-aligned with our needs, since the envisaged benchmark aims to assist the evaluation of methods for sensational image detection in news items.

\subsection{Data selection}
\label{sec:data_selection}

\begin{figure*}[t]
    \centering
    \includegraphics[width=\textwidth]{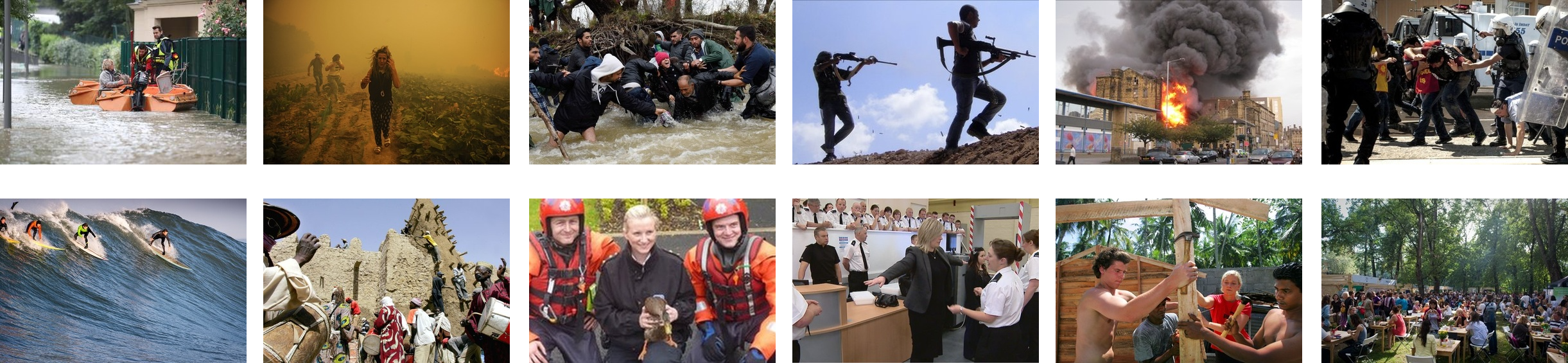}
    \caption{Examples of sensational (top row) and non-sensational (bottom row) images from the created Sens-VisualNews dataset.}
    \label{fig:examples}
\end{figure*}

Candidate ``sensational'' image selection was based on a list of sensational visual concepts and events, that typically appear in disinformation items. This list was created by a team of journalists and fact-checkers that contribute to a project which builds AI-based solutions for automated monitoring of social and news media, to facilitate the timely identification of check-worthy content and emerging disinformation items. This team was formed by $5$ journalists and $7$ fact-checkers from $6$ institutions located in $4$ different countries. The created list contains $194$ sensational visual concepts and events, usually found in media items from the following news topics: i) war \& conflict, ii) refugees \& migrants, iii) climate change \& environmental disasters, iv) racism, and v) religion. 

The automated selection of an initial subset of images from the VisualNews dataset, based on the aforementioned list, was made using a cross-modal network architecture for ad-hoc text-driven video search \cite{Galanopoulos_2025_CVPR} (after making the necessary adaptations for performing large-scale image–text similarity estimation). The employed network architecture computes the semantic similarity between the visual content of an image and the textual description of each item in the created list, using multiple pre-trained VLMs (e.g., CLIP \cite{radford2021learning} and SigLip \cite{zhai2023sigmoid}) and a fixed-weight combination across the utilized VLM families. So, it associates each image of the VisualNews dataset with $194$ similarity scores (one per different sensational visual concept or event from the list).

After retaining the highest score per image and ranking them according to these scores, we selected the $5,000$ high-scoring images as candidates of the ``sensational'' class. However, a statistical analysis of the associated news topics indicated the under-representation of specific topics. In particular, $87\%$ of the selected images were associated with the ``war \& conflict'' and ``refugees \& migrants'' news topics. To mitigate any bias in the image selection process, an additional set of $3,000$ images from the under-represented news topics, was selected from VisualNews by taking into account both the assigned similarity score by the employed network architecture, and the associated sensational visual concept or event. Following, an equivalently large set of candidate images for the ``non-sensational'' class was formed by choosing the $5,000$ images with the lowest similarity scores, and selecting $3,000$ images more using the same network architecture and a curated list of visually-relevant, non-sensational concepts and events (e.g., we used ``families in camping tents'' as an adversary to ``people in refugee camps''). Finally, duplicate images were removed by comparing image hashes. 

\subsection{Data annotation}

\begin{figure}[t]
\centering
\includegraphics[width=0.72\columnwidth]{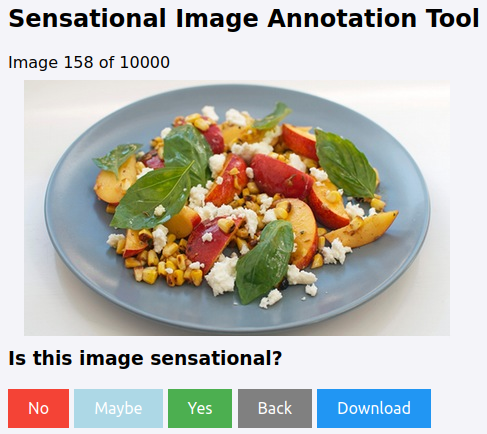}
\caption{A screenshot of the custom-made data annotation tool.}
\label{fig:tool}
\end{figure}

All the selected images were subsequently human-annotated, based on the existence or not of sensational visual content. The annotation was conducted by three annotators using a custom-made web-based annotation tool (see Fig. \ref{fig:tool}). The images were presented on a one-by-one basis and each annotator was asked to specify whether their visual content was sensational, non-sensational or ambiguous, by selecting the relevant button in the user-interface (UI) of the annotation tool. The annotators were able to go back and change their initial decision about an image if need be, and they could download a JSON file with the assigned labels after the completion of the annotation process. The three annotators worked independently and without any knowledge about the choices of each other; each annotator processed the entire set of selected images. Upon completion of the annotation process, the three individual labels that were collected for each image, were aggregated based on majority voting. In case of a tie or an ``ambiguous'' decision, the image was not included in the dataset, as this lack of consensus indicated that the assessment of the image was too subjective. Through the annotation process, $24\%$ of the candidate ``sensational'' images were flipped to the ``non-sensational'' class. This statistic indicates the effectiveness of the adopted method for automated data selection.

\begin{figure}[t]
\centering
\begin{tikzpicture}
\small
\pie[rotate=0, text=legend, radius=2]{
32.0/{War \& Conflict},
29.3/{Migrants \& Refugees},
18.4/{Environ. Disasters},
14.5/Racism,
1.8/Religion,
4/Other
}
\end{tikzpicture}
\caption{News topics distribution in the ``sensational'' class of the created Sens-VisualNews dataset.}
\label{fig:concepts}
\end{figure}
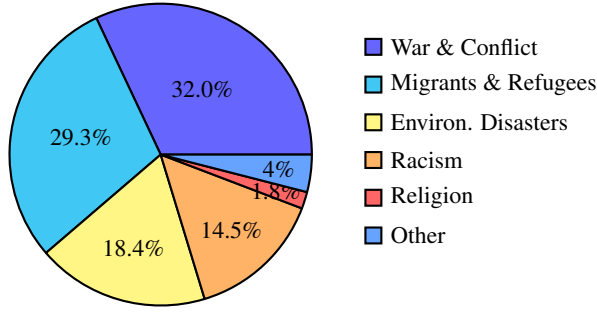

The created Sens-VisualNews dataset through the conducted data annotation process, includes 9,576 images that are equally divided in two different classes (``sensational'' / ``non-sensational''). Indicative examples of images from these two classes are shown in Fig. \ref{fig:examples}. Based on the applied two-step data selection strategy, the created Sens-VisualNews dataset contains diverse visual content from various news topics, as demonstrated by the pie chart in Fig. \ref{fig:concepts}. 

\subsection{Strict subset}

The threshold for a visually-triggered emotional response varies significantly across humans; thus, there is an inherent subjectivity in the task of classifying an image as sensational or not. As a consequence, the created Sens-VisualNews dataset can be susceptible to label ambiguity. To address this concern, we formed a smaller variant of the dataset (called strict subset from now on) by keeping only a subset of images of the ``sensational'' class, for which a strict unanimous consensus was recorded across the three annotators, and choosing an equivalent number of images from the ``non-sensational'' class. The formed strict subset offers a more robust performance evaluation on the sensational image detection task, since it is less affected by label ambiguity. However, this comes at the cost of reducing the complexity of the task, since the more nuanced and challenging images are excluded. For this reason, we release both the full dataset and the strict subset of it, aiming to encourage further research into the subjective interpretation of such terms.

\section{Experiments}

\subsection{Evaluation protocol}

The performance of various SotA open MLLMs on the introduced task of sensational image detection, was evaluated based on top-1 accuracy, after comparing each model's response (generated with greedy decoding) with the corresponding ground-truth label for each sample of the test set. Moreover, we considered two different evaluation settings: in the \textbf{zero-shot setting}, each model is used as-is without any further training on the task; in the \textbf{fine-tuned} setting, the models showing the highest zero-shot performance are adapted to the task using a subset of images (called development set from now on) that contains approx. $10\%$ of the samples from each class of the full Sens-VisualNews dataset. Fine-tuning was performed using LoRA (Low-Rank Adaptation) \cite{hu2022lora} for $30$ epochs with $3$ epochs for warmup. We set $r$ equal to $16$, $\alpha$ equal to $32$, batch size equal to $64$, and learning rate equal to $2\cdot10^{-4}$ and updated based on a cosine schedule.

\subsection{Used MLLMs and prompting strategies}

We took into account the following open MLLMs that exhibit SotA benchmark results (the numbers in parenthesis indicate the amount of parameters in Billions): i) Qwen3 VL \cite{Qwen3-VL} (2B, 4B, 8B), ii) LLaVA OneVision \cite{li2024llava} (0.5B, 7B), iii) LLaVA OneVision 1.5 \cite{an2025llava} (4B, 8B), iv) InternVL 3.5 \cite{wang2025internvl3} (1B, 2B, 4B, 8B), and v) SmolVLM2 \cite{marafioti2025smolvlm} (2.2B). Following, since the term ``sensational image'' is inherently subjective, we considered it necessary to provide a definition of it or describe specific features of such an image, when prompting the MLLMs. Moreover, to examine the sensitivity of the aforementioned MLLMs to the used prompt, we examined $3$ different prompts and $2$ prompt formats for each prompt (prepending or appending the visual tokens in the prompt - if supported by the model) and ran experiments on the development sets of the full Sens-VisualNews dataset and the strict subset of it. The used prompts were the following:
\begin{enumerate}
\item \textit{``Is this image sensational? A sensational image evokes strong emotions (e.g. fear, anger, anxiety, disgust, shock). Answer with a single yes or no.''}
\item \textit{``Does this image contain shocking, provocative or emotionally charged features to grab the viewer's attention? Answer with a single yes or no.''}
\item \textit{``Does this image trigger strong emotional responses (e.g. fear, anger, anxiety, disgust, shock)? Answer with a single yes or no.''}
\end{enumerate}

\subsection{Experimental results}

The results of the conduced prompt sensitivity analysis are presented in Table~\ref{tab:prelim}. The models' sensitivity is quantified by computing the mean and standard deviation of the top-1 accuracy across the different runs. The gradually higher scores of the mean top-1 accuracy for models of the same family as the number of parameters increases, in both of the development sets, show a relation between the model's understanding capacity and detection accuracy. On the contrary, there is no strict connection between the model's size and prompt sensitivity. Nevertheless, the computed standard deviation scores for both of the development sets indicate a noticeable sensitivity for all the different models, which can be extremely high for very small models, such as LLaVA OV 0.5B.

\begin{table}[t]
\centering
\caption{Prompt sensitivity analysis using the two development sets. For each model, we present the mean and standard deviation for the top-1 accuracy measure.}
\label{tab:prelim}
\vspace{7pt}
\resizebox{\columnwidth}{!}{%
\begin{tabular}{lccc}
\toprule
& & \textbf{Dev} & \textbf{Dev} \\
\textbf{Model} & \textbf{Params} & \textbf{Strict} & \textbf{Full} \\
\midrule
SmolVLM2 \cite{marafioti2025smolvlm} & 2.2B & $64.4 \pm 5.9$ & $59.2 \pm 4.2$ \\
\midrule
\multirow{4}{*}{InternVL 3.5 \cite{wang2025internvl3}} & 1B & $74.8 \pm 1.2$ & $71.6 \pm 3.8$ \\
& 2B & $81.9 \pm 5.0$ & $74.0 \pm 4.8$ \\
& 4B & $82.8 \pm 3.3$ & $75.0 \pm 4.7$ \\
& 8B & $89.9 \pm 3.1$ & $80.1 \pm 3.0$ \\
\midrule
\multirow{3}{*}{Qwen3 VL \cite{Qwen3-VL}} & 2B & $86.0 \pm 5.9$ & $81.8 \pm 4.4$ \\
& 4B & $86.6 \pm 2.0$ & $83.0 \pm 1.7$ \\
& 8B & $88.1 \pm 4.2$ & $83.2 \pm 3.0$ \\
\midrule
\multirow{2}{*}{LLaVA OV \cite{li2024llava}} & 0.5B & $73.1 \pm 15.8$ & $67.4 \pm 13.2$ \\
& 7B & $90.7 \pm 1.8$ & $80.8 \pm 2.2$ \\
\midrule
\multirow{2}{*}{LLaVA OV 1.5 \cite{an2025llava}} & 4B & $90.1 \pm 1.6$ & $83.1 \pm 2.4$ \\
& 8B & \boldmath $90.3 \pm 3.7$ & \boldmath $83.4 \pm 3.1$ \\
\bottomrule
\end{tabular}}
\end{table}

Further evaluation of the MLLMs was made on the test set of full Sens-VisualNews dataset and its strict subset. For each MLLM, we used the prompt and prompt format that led to the best performance on the development set. In this evaluation, we also took into account a baseline approach; i.e., a variant of the employed VLM-based approach for data selection (see Section \ref{sec:data_selection}), that takes a final decision about an image based on a threshold. Images with similarity scores above/below the selected threshold are classified as sensational/non-sensational. This threshold was determined based on the development set and was set equal to the median relevance score, reflecting the balanced class distribution. Finally, as stated before we evaluated all models' performance on a zero-shot setting, and we further assessed the best-performing ones after fine-tuning them.

The results about the performance of the baseline VLM-based approach and the various MLLMs on the tests sets of the full Sens-VisualNews dataset and its strict subset, are reported in Table \ref{tab:results}. The zero-shot performance of MLLMs on the test set of the full dataset is closely tied to the total number of model parameters (as expected), while a similar trend is observed for most of the models in the strict subset. Moreover, the observed performance on the strict subset is consistently higher compared to the full dataset (as also expected), due to the decreased label ambiguity/noise. In addition, most of the examined MLLMs outperform the baseline approach by a noticeable margin, despite the fact that this approach exhibits more than $80\%$ top-1 accuracy on both test sets. Finally, the highest zero-shot performance on both test sets is observed for the LLaVA OV 1.5 8B model, which performs slightly better than the equivalently large model from the Qwen3 VL family.

Driven by the findings reported above, we focused on two families of models (LLaVA OV 1.5 and Qwen3 VL) and examined their performance after fine-tuning them on the development set of the full Sens-VisualNews dataset. As can be seen in the gray-coloured rows of Table \ref{tab:results}, the fine-tuning resulted in significantly improved performance in all cases. Notably, the fine-tuned version of Qwen3 VL 8B gains $+3.0\%$ top-1 accuracy on the full test set compared to the zero-shot version, and exhibits the highest performance among all MLLMs on the Sens-VisualNews dataset. Overall, despite the fact that this task is highly challenging due to its subjectivity, the performance of modern MLLMs on the created Sens-VisualNews dataset demonstrates their advanced competence to spot sensational images and assist the detection and flagging of potential disinformation.

\begin{table}[t]
\centering
\caption{Performance comparison on the full Sens-VisualNews dataset and its strict subset, across zero-shot and fine-tuned settings. For each setting: Best scores bold, second-best underline. Gray highlight indicates fine-tuning.}
\label{tab:results}
\vspace{7pt}
\begin{tabular}{lccc}
\toprule
& & \multicolumn{2}{c}{\textbf{Test Top-1 (\%)}} \\
\textbf{Model} & \textbf{Params} & \textbf{Strict} & \textbf{Full} \\
\midrule
Baseline (VLM-based) & 11B & 84.2 & 81.2\\
\midrule
SmolVLM2 \cite{marafioti2025smolvlm} & 2.2B & 68.1 & 62.2 \\
\midrule
\multirow{4}{*}{InternVL 3.5 \cite{wang2025internvl3}} & 1B & 76.4 & 76.1 \\
& 2B & 86.2 & 77.9 \\
& 4B & 88.8 & 83.4 \\
& 8B & 92.1 & 83.8 \\
\midrule
\multirow{6}{*}{Qwen3 VL \cite{Qwen3-VL}} & 2B & 92.7 & 86.3 \\
& 4B & 90.8 & 85.0 \\
& 8B & \underline{92.8} & \underline{87.0} \\
& \cellcolor{lgray} 2B (ft) & \cellcolor{lgray}95.1 & \cellcolor{lgray}89.4 \\
& \cellcolor{lgray} 4B (ft) & \cellcolor{lgray}\underline{95.4} & \cellcolor{lgray}89.4 \\
& \cellcolor{lgray} 8B (ft) & \cellcolor{lgray}\textbf{95.5} & \cellcolor{lgray}\textbf{90.0} \\
\midrule
\multirow{2}{*}{LLaVA OV \cite{li2024llava}} & 0.5B & 90.8 & 81.3 \\
& 7B & 91.8 & 81.5 \\
\midrule
\multirow{4}{*}{LLaVA OV 1.5 \cite{an2025llava}} & 4B & 89.7 & 86.7 \\
& 8B & \textbf{93.9} & \textbf{87.6} \\
& \cellcolor{lgray}4B (ft) & \cellcolor{lgray}\underline{95.4} & \cellcolor{lgray}89.7 \\
& \cellcolor{lgray}8B (ft) & \cellcolor{lgray}94.6 & \cellcolor{lgray}\underline{89.8} \\
\bottomrule
\end{tabular}
\end{table}

\section{Conclusions}

In this work, we introduced the task of sensational image detection that aims to spot images which trigger strong emotional responses, and proposed the Sens-VisualNews benchmark dataset with 9,576 news images annotated based on the (in-)existence of various sensational visual concepts and events. Using Sens-VisualNews, we studied the sensitivity of various SotA MLLMs on the used prompt, and assessed their performance and robustness on this task, across zero-shot and fine-tuned settings. Our experimental evaluations demonstrated the competency of modern open MLLMs from the Qwen3 VL and LLaVA OV 1.5 families, forming the ground for future comparisons on sensational image detection.

\vfill\pagebreak


\end{document}